# Travel Speed Prediction with a Hierarchical Convolutional Neural Network and Long Short-Term Memory Model Framework

Wei Wang[1] and Xucheng Li[2]

[1] Atkins(SNC-Lavalin), UK,
*w.wang@atkinsglobal.com*

[2] Shenzhen Urban Transport Planning Center Co. Ltd, China,
*xucheng.li@sutpc.com*

**Abstract**

Advanced travel information and warning, if provided accurately, can help road users avoid traffic congestion through dynamic route planning and behavior change. It also enables traffic control centres mitigate the impact of congestion by activating Intelligent Transport System (ITS) proactively. Deep learning has become increasingly popular in recent years, following a surge of innovative GPU technology, high-resolution, big datasets and thriving machine learning algorithms. However, there are few examples exploiting this emerging technology to develop applications for traffic prediction. This is largely due to the difficulty in capturing random, seasonal, non-linear, and spatio-temporal correlated nature of traffic data. In this paper, we propose a data-driven modelling approach with a novel hierarchical D-CLSTM-t deep learning model for short-term traffic speed prediction, a framework combined with convolutional neural network (CNN) and long short-term memory (LSTM) models. A deep CNN model is employed to learn the spatio-temporal traffic patterns of the input graphs, which are then fed into a deep LSTM model for sequence learning. To capture traffic seasonal variations, time of the day and day of the week indicators are fused with trained features. The model is trained end-to-end to predict travel speed in 15 to 90 minutes in the future. We compare the model performance against other baseline models including CNN, LGBM, LSTM, and traditional speed-flow curves. Experiment results show that the D-CLSTM-t outperforms other models considerably. Model tests show that speed upstream also responds sensibly to a sudden accident occurring downstream. Our D-CLSTM-t model framework is also highly scalable for future extension such as for network-wide traffic prediction, which can also be improved by including additional features such as weather, long term seasonality and accident information.

# 1. Introduction

Over the last 20 years, the vehicle miles for all road types have increased by 15.8%, with 32.6% for motorways, 23.9% for rural 'A' roads and 20.9% for rural minor roads in the UK (Department for Transport, 2018). This increasing amount of traffic is putting more and more pressures on road infrastructures, and it is currently estimated that traffic congestion costs motorists over £37.7 billion in 2017, with drivers spending an average of 31 hours a year in congestions during peak hours. This makes the UK one of the top ten most congested countries in the world (Inrix, 2017).

To ease congestions and mitigate the costs from congestions on UK motorways, Smart Motorway uses Active Traffic Management techniques and applies variable speed limits and hard shoulder running. For urban areas, Urban Traffic Management and Control (UTMC) systems were developed to tackle congestions and proactively prevent or reduce the effect of congestions. Advanced traveller information and warning, if provided timely and accurately, can help road users avoid traffic congestion through advanced route planning and behaviour change including mode shift. The success of these Intelligent Transport System (ITS) applications rely heavily on traffic condition prediction.

Traffic prediction is regarded as one of the most challenging tasks in transport planning sector, even with the aid of big datasets such as mobile network data and global position system (GPS) and the evolution of computer processing power. This is largely due to the random, seasonal, non-linear, and spatio-temporal correlated nature of traffic data.

Travel patterns of flows and speeds generally show strong seasonal and periodic variations, which are difficult to be captured by the traditional empirical models. In congested travel condition, traffic queues related to incidents or unexpected events propagate rapidly over the time and space domain. Meanwhile, motorway shockwaves, caused by unexpected lane changing or sudden vehicle brake, occur randomly and their dissipated rates vary significantly, which are difficult to predict. Furthermore, a network-wide traffic prediction is even more difficult due to the variations of local driving behaviour, traffic composition and temporary roadworks, network structure and facility.

In this paper, a novel hierarchical deep learning model, D-CLSTM-t, is proposed for short-term traffic speed prediction, a framework leveraging the rapidly-developed convolutional neural network (CNN) and long short-term memory (LSTM) model technique. Our contributions to knowledge in this paper are summarised as follows:

- A deep learning end-to-end model (D-CLSTM-t) is proposed to predict traffic speed, which combines a 2-dimensional CNN with space and time information built on the same graph to capture latent traffic patterns and subsequent LSTM blocks to learn the sequence features of traffic data. By varying different filter size and introducing a Dual CNN-LSTM architecture and a time marker layer, this model is trained successfully to capture the spatial and temporal correlations at the same time, as well as the complex cyclical and seasonal characteristics of traffic data.
- Based on historical traffic flow and speed from the previous time steps on a selected motorway section on M25 in the UK, our model can predict travel speed for all locations simultaneously in the next 15 to 90 minutes. With the proposed model structure, our model outperforms other baseline models and methods based on real traffic data inputs from a selected section on M25 in the UK. It can be used to provide network-wide real-time traffic prediction and evaluate queue propagation during traffic events. Model tests show that speed upstream also responds sensibly to a sudden accident occurrence at downstream when flow throughput is significantly reduced.
- The model developed in this study is easy to train and less computational expensive. The model is developed with an architecture to fuse a wide range of data inputs including historical speed, flow and seasonality information, which guarantees the scalability of the model to introduce other factors such as weather and site-specific information. It can also be extended to a network wide traffic prediction and has potential for transfer learning to model other graphic spatio-temporal and time series problems.

The rest of this paper is arranged as follows. Section 2 reviews the related work and existing literature for short-term traffic prediction. Section 3 briefs the deep learning architecture for CNN and LSTM. Out model methodology and the data used for the model development are presented in section 4. The model output comparisons are detailed in section 5 including a speed prediction in response to a simulated accident by using the trained D-CLSTM-t model. At the end this paper, we summarise our findings and some future work extension is discussed.

## 2. Related Work

In past decades, many research activities and attentions has been devoted to the theme of predict short-term traffic from historical datasets. In earlier days, the method of some of the work was primarily focused on the development of statistical and mathematical models to capture the relationships between traffic parameters such as flow, speed, density and headway etc.

Traffic data has time series nature and early studies have applied statistical approaches such support vector machines (SVM) and the auto regressive integrated moving average (ARIMA) for traffic prediction (Wu and Lee, 2004; Williams and Hoel, 2003). However, these techniques are typically not successful to capture the spatio-temporal relationships of traffic data in a wide transport network.

Recently, with emerging learning technique and big datasets, the study of machine learning and deep learning in the transport sector becomes increasingly popular. A comprehensive literature review on short-term traffic forecast was provided by Vlahogianni et al. (2014). Du et al. (2018) developed a hybrid multimodal deep learning framework (HMDLF) for traffic flow forecasting, which consists of one dimensional convolutional neural network (CNN) and gated recurrent units (GRU). Inspired by early work in the combination of convolution and recurrent neural network (Shi et al., 2015), Wu and Tan (2016) developed a hybrid one-dimensional CNN and LSTM sequential model to predict traffic flows, taking account of short-term temporal and periodic traffic features. Polson and Sokolov (2016) developed a deep neural network model to predict traffic flows during a special event and unusual weather condition. To predict citywide crowd flows, Zhang et al. (2017) designed a Deep Spatio-Temporal Residua Network model (St-ResNet) to capture temporal closeness, period and trend properties of crowd traffic, supported by the introductions of external factors such as weather and public events.

The deployment of CNN for a network-wide traffic speed was proposed by Ma et al. (2017) to transform traffic speed into a series of image graphs, which outperformed other state-of-art models. Inspired by the approach, a recent study carried out by Kim et al. (2018) developed a pioneering neural network with capsules (Hinton et al., 2018) technique that replaces max pooling in CNN layers by dynamic routing, which outperforms the state-of-the-art method of a traditional CNN model by 8.6% measured root mean squared, however with an expense of longer model training time compared with CNN. A study by Yu et al. (2017) was attempted to predict network-wide traffic by spatiotemporal recurrent convolutional networks (SRCNs), with the CNN to capture the near and far-side spatial dependencies and LSTM to learn the long-term temporal patterns. The model was trained on images converted from a series of network-wide speed plots on small grids.

Most previous studies have focused on short-term traffic flow prediction. The prediction of travel speed is more difficult especially when the primary input data available is based on historical traffic flows, largely due to the non-linear relationship between flow and speed, as shown in **Figure 1**. It clearly shows a two-value speed-flow curve where a given flow can be associated with two different speeds, converging to an optimum status when flow reaches road capacity. A low flow value may be related to a free flow travel condition such as in the off-peak when traffic is light, but it may also correspond to traffic congestion in peak periods when the flow throughput is heavily reduced. Traffic data also has a strong time of the day and seasonal characteristics, as discussed in more detail later. Based on a spatio-temporal graph convolutional neural network (ST-GCNN) combined with gated linear units(GLU) and highway network, Yu et al. (2017) achieved a state-of-the-art model performance in terms of traffic speed prediction against two sets of historical speed datasets.

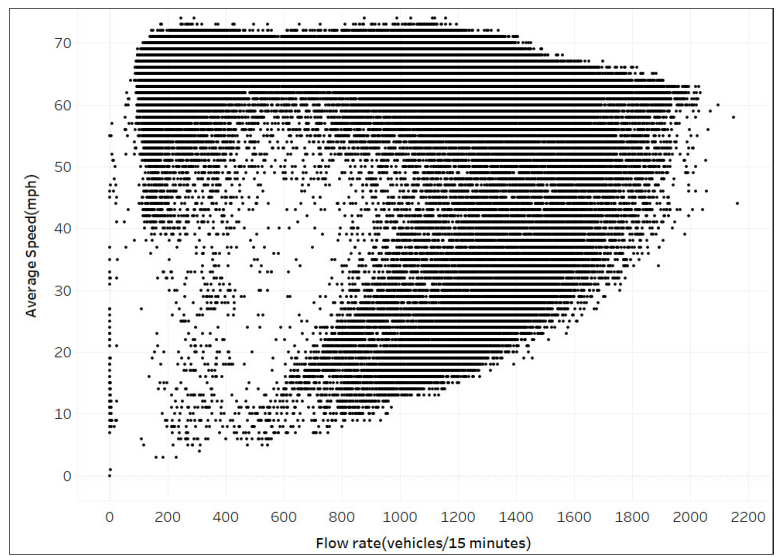

**Figure 1 Flow-speed scatter plot, M25 CW between J8 to J12, weekdays, September 2017**

**3.     Deep Learning Architecture**

**3.1     Convolutional Neural Network (CNN)**

For its innovative method of extracting latent feature patterns from images, the CNN (LeCun, and Bengio, 1995) is one of the structures in the Neural Network family that proved to have standout performance in multiple challenging tasks in computer vision, including object detection, image recognition, video recognition, text detection etc.

A typical CNN architecture, as shown in **Figure 2**, is formed by components of a convolutional layer, a pooling layer, a fully connected layer. The convolutional layer is the first layer of a CNN model, which applies a convolutional kernel as filter (normally a x by x matrix) to calculate the weights for each local area (size depending on the filter size). The output becomes so call feature maps, which highlights local features by grouping adjacent input values with the designed convolutional kernels. The convolutional kernels could be applied to multiple input channels to dig out more generalised features from the input dataset. Pooling layers come after the convolutional layers, which combine the neuron outputs into one single neuron. By doing so, the model structure further clusters the features and the characteristics hidden in the given information. Typical pooling methods are maximum pooling and average pooling, which apply the maximum values or the average values in the sliding windows respectively. The output matrices from the pooling layers are flattened to become the fully connected layers, which applies same calculation mechanism as traditional artificial neural network (ANN). Multiple convolutional and pooling layers can be stacked to increase the depth of the model to maximise its ability in feature learning.

One of advantages of CNNs is to reduce computation time and memory consumption by sharing the same filter for hidden neurons. This also reduces the total number of hyperparameters to be trained comparting to the traditional ANN since a set of bias and weights are applied across all receptive fields.  This, however, is one of the drawbacks of CNN that the features learned are location-invariant, or equivariant. This limitation would have a significant impact on traffic prediction, which is highly site specific. On a motorway corridor, for instance, a similar flow on two independent sections at a specific time slot may have a completely different travel speed, especially when some site related features are not specially presented in the input data, as such as ramp merging and diverging, local driving behaviour, slow moving traffic composition etc.  A few recent studies have tried to tackle the issue (Shi et al.,2017; Lin et al., 2018) and significantly improved model performance has been reported against several of benchmark models.

With a series of convolutional layers and pooling layers, CNN structure pre-process localised features from the input. For traffic prediction problems, with input matrices as spatio-temporal diagram, the designed filters will convolve features from adjacent values, such as next a couple of links and next time periods.

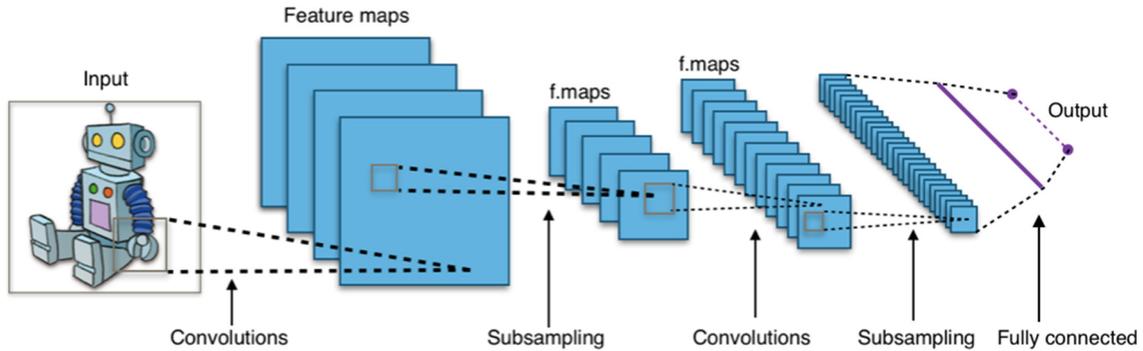

**Figure 2 A typical CNN architecture**
**(image source: https://en.wikipedia.org/wiki/Convolutional_neural_network)**

### 3.2     Long Short-Term Memory (LSTM)

LSTM (Hochreiter and Schmidhuber, 1997) is a special Recurrent Neural Network (RNN) module developed in tackling the knowing gradient vanishing issue for RNN. In contrast to a single layer of a standard RNN, the LSTM is capable of learning long-terms dependencies across samples by employing a number of gates to control the information generated to flow through designed memory cells. These gates include input gate, forget gate and output gate, denoted by it, $f_t$ and $o_t$ respectively.

The architecture of LSTM is illustrated in **Figure 3**.

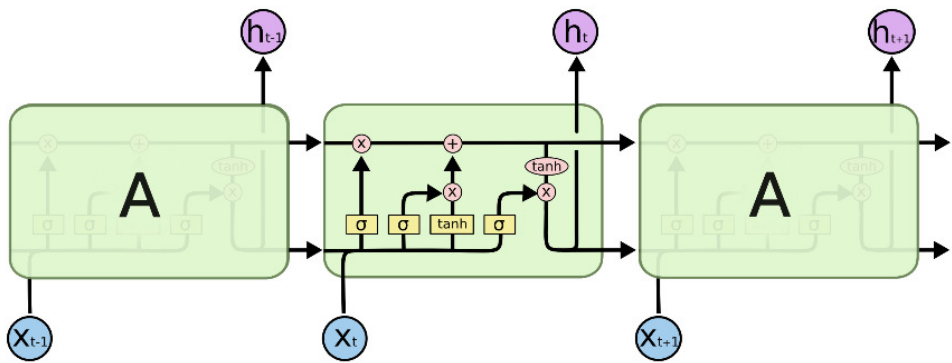

**Figure 3   LSTM layer structure**
**(image source: https://colah.github.io/posts/2015-08-Understanding-LSTMs/)**

The computation of each gate is given as follows:

$$i_t = \sigma(W_i x_t + U_i h_{t-1} + b_i)$$
$$f_t = \sigma(W_f x_t + U_f h_{t-1} + b_f)$$
$$o_t = \sigma(W_o x_t + U_o h_{t-1} + b_o)$$

Where the $W$ and $U$ are the weights for the input vector $x_t$ and previous short-term state $h_{t-1}$ to be learned through training, and the $b$ stands for the bias vectors for each function. $\otimes$ denotes pointwise multiplication operation.

For each LSTM hidden layer, the gate values will be computed, which are then applied to calculate three key states in current layer. The three key states include the potential new candidate values $g_t$ based on the hidden states of previous steps, the updated states $c_t$ as the output of current layer and the hidden state $h_t$ of current step which is not included in the output.

$$g_t = tanh(W_g x_t + U_g h_{t-1} + b_g)$$
$$c_t = f_t \otimes c_{t-1} + i_t \otimes g_t$$
$$y_t = h_t = o_t \otimes tanh(c_t)$$

RNN has been applied successfully for many sequence based tasks such as time series forecasting, speech recognition and sentiment analysis. However, it can only learn spatial or temporal dependencies one at a time. RNN also has a known issue of error accumulation due to feeding model output back into itself (Holden et al., 2017). RNN turns to overfit the training data and generalise poorly (Yu et al., 2017).

## 4. Methodology

### 4.1 Model Architecture

Our proposed model architecture of Dual CNN-LSTM with time marker (D-CLSTM-t) is illustrated in **Figure 4**. The model generally follows the CNN-LSTM structure which has been successful in image captioning, object detection, video description, and semantic segregation tasks (Donahue et al., 2016). The CNN-LSTM framework exploits CNN models for feature leaning and extraction which are then fed into the recurrent LSTM model for interpreting the sequence features across multiple time steps.

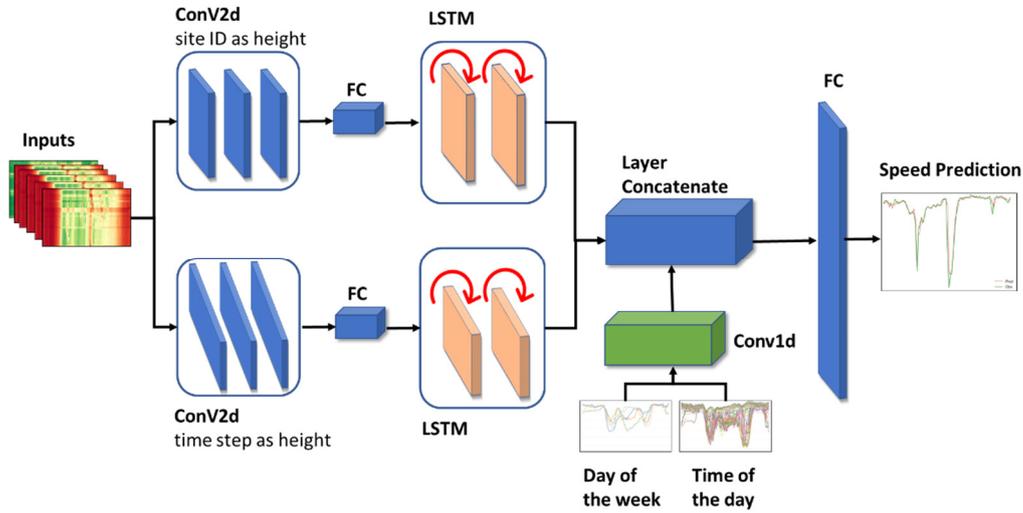

**Figure 4 D-CLSTM-t model architecture**

The core part of our model is to employ two parallel CNN+LSTM frameworks to capture the complexity and richness of traffic characteristics. The key difference between the two is in the sequence training by LSTM, with one focusing on the space and the other on the temporal features. The three CNN layers in both frameworks are implemented in sequence followed by a time distributed fully connected layer, to maintain the integrity of space and time information for the learned position-dependent features before being fed to two LSTM layers for further sequence feature learning.

The outputs from the CNN-LSTM modules are then concatenated with a 1-dimensional convolutional layer (Conv1D) which is deployed to capture time information including time of the day and day of the week. We have investigated different input data format and found that the combination of traffic flows by vehicle length, travel speed, time of the day and day of the week information achieved the best model performance.

For the CNN-LSTM building block, depth-wise separable 2D convolution layers (Chollet, 2017) was employed, which performs spatial convolution feature learning for each input data channel independently followed by a pointwise convolution (1 x 1 filter) that combines all the features learned into the output channels. Separable convolutions are regarded as a way to split a convolution filter into two smaller filters, which is assumed as an extreme version of an Inception module (Szegedy et al., 2015). The separable convolution layer has an advantage of learning richer features with less trained parameters[1], therefore it can reduce model training time substantial and avoid model over-fitting issue. Our model experiment shows that in general the total number of trained parameters is less than half of these required in a traditional convolutional layer. The hypothesis behind SeparableConv2D is that that extracting cross-channels and spatial correlations in the feature maps completely decoupled.

## 4.2  Model Parameters

We train the model end-to-end in Keras (https://keras.io/) with Tensorflow backend. The detail of our D-CLSTM-t model parameter settings is shown in **Table 1**. The model has a total of 570,275 training hyperparameters. The activation function of ReLU (Rectified Linear Unite) is applied in the convolutional layer, whereas the function of tanh is applied in the LSTM layers. Each convolution layer is followed by a batch normalisation layer to accelerate training by reducing internal covariance shift. Note that varied filter is introduced across the convolutional layers to capture the complex spatio-temporal traffic patterns. Meanwhile, since the data image input are relatively small, no pooling layers are employed in the convolutional layers, which are consistent with the evidence found in other studies (Springenberg et al., 2014, Wu and Tan, 2016).

Adam (Adaptive Moment Estimation) is selected as the optimizer of the model, and l2 norm regularisation on weights is applied to prevent overfitting for the final regression dense layer. A grid-search process was conducted to find the best values for the three important hyperparameters that have significant influences on out model performance, namely, the learning rate (lr) for Adam optimizer, λ for l2 norm, and the batch size for gradient decent optimisation, and the selected values are 0.003, 0.0002, and 4, respectively.

A maximum of 50 epochs is set up for model fitting on training and validation datasets, and 'early stopping' is activated to avoid over-fitting on training data. The mean squared errors (MSE) between model output and the expected output are calculated as loss function, while the mean absolute errors (MAE) are used as the complementary measures for model performance evaluation, defined as follows.

$$MSE = \frac{1}{n_p} \sum_{t=1, s=1}^{n,m} (y_{ts} - \widehat{y_{ts}})^2$$

---

[1] For example, considering an input layer with a size of (batch, 60, 4, 6), kernel size (2,1) and an output channel 32. The separable conv2D firstly implements a channel wise calculation with a total parameter of 2*1*6 = 12, followed by a layer concatenate. Then a depth-wise convolution is applied with a kernel size of (1, 1), with a total of parameter 1*1*6*32 = 192. The total parameters of separable conv2D is 12+ 192 + 32(bias term) = 236, in contrast to a total parameter of 2*1*6*32+32 = 416 for a traditional convolutional layer.

$$MAE = \frac{1}{n_p} \sum_{t=1, s=1}^{n,m} |y_{ts} - \widehat{y_{ts}}|$$

Where $n_p$ is the total number of predictions, t is the future prediction time step reference and s is the site id number, $y_{ts}$ and $\widehat{y_{ts}}$ are the observed and model predicted travel speed respectively.

| Layer | Name | Input / Output Shape | Filter size (Conv) | Number of parameters |
|---|---|---|---|---|
| a_0 | Main_input_s | (None, 60, 4, 6) | | 0 |
| a_1 | SeparableConv2D (relu) | (None, 60, 4, 32) | (2 ,1) | 236 |
| a_2 | BatchNormalization | (None, 60, 4, 32) | | 128 |
| a_3 | SeparableConv2D (relu) | (None, 60, 4, 64) | (4,2) | 2368 |
| a_4 | BatchNormalization | (None, 60, 4, 64) | | 256 |
| a_5 | SeparableConv2D (relu) | (None, 60, 4, 128) | (8,4) | 10368 |
| a_6 | BatchNormalization | (None, 60, 4, 128) | | 512 |
| a_7 | TimeDistributed_Flatten | (None, 60, 512) | | 0 |
| a_8 | TimeDistributed_Dense | (None, 60, 6) | | 3078 |
| a_9 | LSTM (tanh) | (None, 60, 60) | | 16080 |
| a_10 | LSTM (tanh) | (None, 60, 6) | | 1608 |
| b_0 | Main_input_t | (None, 4, 60, 6) | | 0 |
| b_1 | SeparableConv2D (relu) | (None, 4, 60, 32) | (2,4) | 272 |
| b_2 | BatchNormalization | (None, 4, 60, 32) | | 128 |
| b_3 | SeparableConv2D (relu) | (None, 4, 60, 64) | (2,8) | 2624 |
| b_4 | BatchNormalization | (None, 4, 60, 64) | | 256 |
| b_5 | SeparableConv2D (relu) | (None, 4, 60, 128) | (4,16) | 12416 |
| b_6 | BatchNormalization | (None, 4, 60, 128) | | 512 |
| b_7 | TimeDistributed_Flatten | (None, 4, 7680) | | 0 |
| b_8 | TimeDistributed_Dense | (None, 4, 60) | | 460860 |
| b_9 | LSTM (tanh) | (None, 4, 60) | | 29040 |
| b_10 | LSTM (tanh) | (None, 4, 60) | | 29040 |
| b_11 | Reshaped layer | (None, 60, 4) | | 0 |
| c_0 | TimeMarker_input | (None, 60, 8) | | 0 |
| c_1 | Conv1d | (None, 60, 1) | (60) | 481 |
| d_0 | Concatenate_layer | (None, 60, 11) | | 0 |
| d_1 | TimeDistributed_Dense | (None, 60, 1) | | 12 |
| **Total training parameters** | | | | **570275** |

Table 1 D-CLSTM-t model parameter settings

## 5. Case Study

### 5.1 Dataset

To illustrate the model performance, a case study is undertaken using the traffic data on M25 motorway in the UK. The traffic data for this paper is downloaded from Highways England's WebTRIS website[2], a map-based data query service with seed and flow collected from road sensors along the Strategic Road Network (SRN). The daily report of traffic data for 68 site locations between M25 motorway clockwise section from Junction 8 to junction 12, covering an approximate 30-kilometer road section, as shown in **Figure 5**. The selected M25 section has four lanes with a standard lane width, except for some short sections at J10 and J11 which have only three lanes due to specific lane-drop off-slip arrangement.

---

[2] Reference: http://webtris.highwaysengland.co.uk/

The time period for the data collected for model tests is from 04/09/2017 to 31/10/2017, a typical neutral month period avoiding travel patterns affected by holiday seasons and school half terms. The data is available at 15 minutes interval which includes average speed and traffic flows by four vehicle axle length bands, namely, 0-5.2 meters, 5.2-6.6 meters, 6.6-11.6 meters and 11.6+ meters.

Following a review of the data, eight sites are discarded due to significant levels of missing or incomplete data records. For similar reason, data for all 60 remaining sites between midnight and 1AM is also excluded. Thus, each site preserves 92 data points per day. For modeling simplicity, we only consider data for weekdays, which leaves us with traffic 3,862 data points in total for 42 days. We select 35 days, 5 days and 2 days of data for training, validation and test respectively. The validation dataset covers a randomly selected weekday spectrum to ensure a generalisation of model performance.

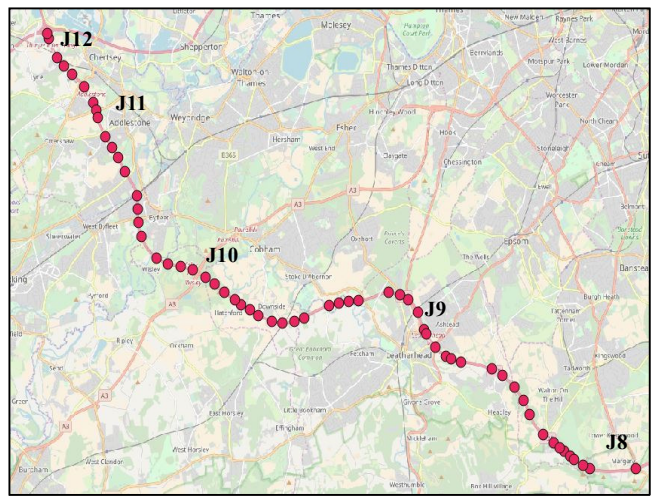

**Figure 5 WebTRIS site location map for this paper**

**Figure 6(a)** and **Figure 6(b)** show day of the week and time of the day variations for travel speed on M25 between J8 and J12. As can be seen from **Figure 6(a)** for the site M25/4717A near M25 J10, there are significant variations for the travel speed distributed across the weekday, reflecting a typical heavier than normal traffic congestion occurring in morning peak period in Monday and Friday afternoon. Meanwhile, there are also widely distributed speed patterns across the day for all 60 sites, as shown in **Figure 6(b)**. These variations, along with the non-linear relationship between flow and speed, factors such as unexpected event, accident, slow moving vehicle, weather etc., bring lots of challenges for robust short-term traffic prediction by traditional linear regression model or static mathematic formulations.

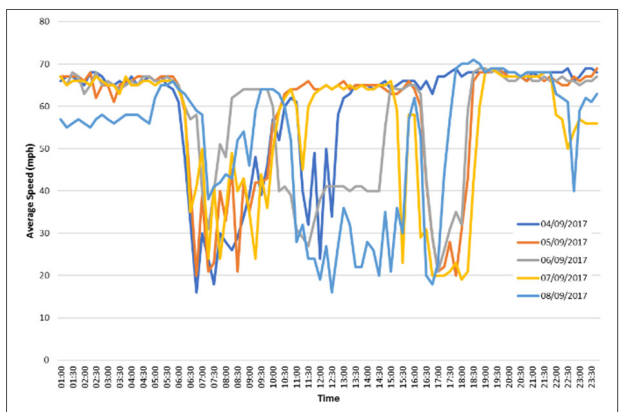
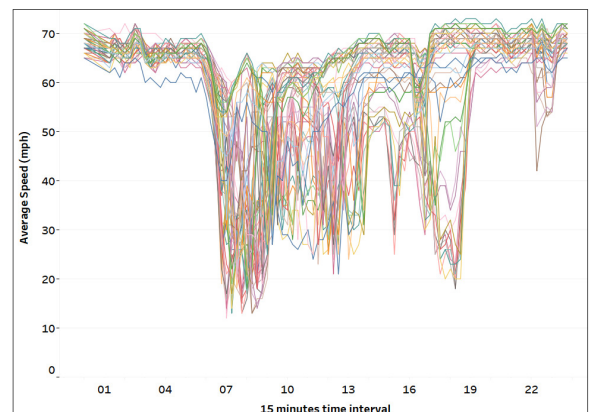

(a) Day of the week variations, site M25/4717A, Monday to Friday from 04/09 to 08/09, 2017

(b) Time of the day variations, all 60 sites between M25 J8 and J12, Monday, 04/09/2017

**Figure 6 Day of the week and time of the day variation for travel speed on M25**

## 5.2 Data processing and setup

For the data points retained, it is found that a few sites have some gaps with no data being recorded. These are infilled by the most recurrent data on the same day if the gap is smaller than 4 continuous time slots. Otherwise the data from the proceeding same weekday is applied to maintain the day of the week characteristics of the traffic data. **Table 2** below shows the rates of missing data by month. Overall, the missing data record is approximate 2.11%.

| Calendar Month | Number of weekdays | Number of raw valid data records | Total expected data records | Missing data rate |
|---|---|---|---|---|
| Sep-17 | 20 | 109363 | 110400 | 0.94% |
| Oct-17 | 22 | 117586 | 121440 | 3.17% |
| Total | 42 | 226949 | 231840 | 2.11% |

**Table 2 Summary of data records**

In this paper, our objective is to use historical traffic data information (t-n, t-n+1, ..., t-1) from p site locations $\{s_i\}_{i=1}^{p}$ to predict site speed $\{y_i\}_{i=1}^{p}$ in the (t, t+1, ..., t+h) in the horizon future. Typically, the parameter of **n** corresponds to the number of lagged observations in the traditional time series model, and the parameter **h** is related to how far the model is capable of predicting future events. In most studies, to achieve desired prediction accuracy **h** is a short-term time window of 15 or 30 minutes in the future.

Our spatio-temporal input matrix is given below:

$$S = \begin{bmatrix} s_{1(t-n)} & s_{1(t-n+1)} & \cdots & s_{1t} \\ s_{2(t-n)} & s_{2(t-n+1)} & \cdots & s_{2t} \\ s_{3(t-n)} & s_{3(t-n+1)} & \cdots & s_{3t} \\ \vdots & \vdots & \cdots & \vdots \\ s_{p(t-n)} & s_{p(t-n+1)} & \cdots & s_{pt} \end{bmatrix} \in \mathbb{R}^{p \times n}$$

The matrix S has a space dimension across p sites and a time dimension n for historical window slices.

Traffic flow and speed are affected by travel condition at downstream over the time. For instance, traffic queue as a result of road accident may propagate to upstream rapidly and take a long period to dissipate even the accident is cleared away. Thus, for our study, the site locations are arranged in the reversed travel direction starting from downstream to upstream. Similarly, the time dimension is ordered from the most past to the most recent time slot. As such, p and n can be interpreted as the height and width of an input image graph for a convolutional layer. This helps to the deep CNN layers to learn the spatio-temporal travel patterns simultaneously, and the generated feature learned to be fed to the subsequent LSTM layers.

On the other hand, we include a total of 6 traffic features as the input channel, including historical flows by four vehicle axle length bands, total flow and travel speed. Consequently, each data point is constructed as a 4-dimensional tensor, with a size of (1, 60, 4 ,6). In our model experiment, we select a past window size n of 4, which means we exploits the last hour flow and speed information to predict travel speed in the next 15 minutes. For a single workday, we create a total of 88 sample data points. The tensor dimensions for the training, validation and test dataset are then (3080, 60, 4, 6), (440, 60, 4, 6) and (176, 60, 4, 6) respectively. As our study is focused on predicting travel speed in the next 15 minutes, the tensor of the target speed **y** is therefore 3-dimensional. For example, the dimension of y for training data is (3080, 60, 1). The input flow datasets are scaled by a standardisation process to have a mean of zero and a standard deviation of one, whereas speed data is simply normalised by dividing each attribute by 100. Our data processing adopts the conceptual structure of a traditional image, but the processed image is relatively small, mainly due to the limited temporal features presented.

**Figure 7(a)** and **Figure 7(b)** show two heatmap snapshots of speed and total flow data on 04/09/2017, with space on vertical axis and time horizonal axis respectively. It is observed that speed is reduced substantially during peak periods especially in AM peak between 6:00 and 9:00 AM. There is a strong correlation between flow and speed. Meanwhile, from the flow graph, there are some clear horizontal strip patterns, which are related to the sites before and after motorway ramps where flows are increased or reduced significantly.

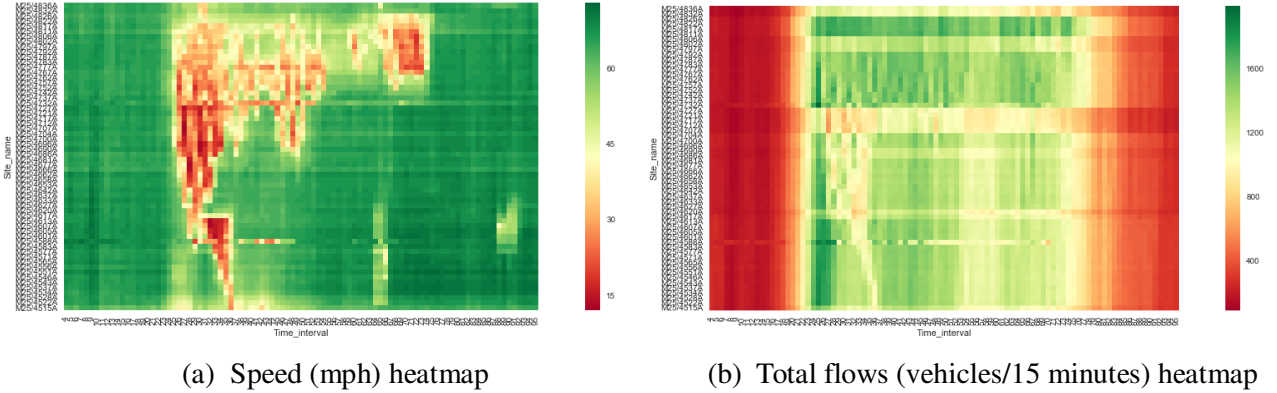

(a) Speed (mph) heatmap        (b) Total flows (vehicles/15 minutes) heatmap

**Figure 7 Speed and total flow time and space heatmap plots on 4<sup>th</sup> September 2017**

## 6. Output Comparison and Analysis

### 6.1 Baseline Model Comparisons

We select a variety of baseline models and forecasting method to evaluate the performance of our D-CLTSM-t (test a) model, including[3]:

- test b: D_CLSTM-t model with traditional convolutional neural network method,
- test c: CLSTM(S)-t model with single CNN+LSTM module, space as graph height,
- test d: CLSTM(T)-t model with single CNN+LSTM module, historical time step as graph height,
- test e: D_CLSTM without time marker,
- test f: CNN-t with only convolutional layers, and
- test g: tree-based model with Light GBM algorithm (Ke et al., 2017).

For model comparison purpose, we also include a speed prediction by common-sense or naïve forecast (test h), i.e. assuming current speed as the prediction for next 15 minutes. And finally, a baseline model of prediction by traditional speed flow curves (test i) is also presented, with the travel time calculated by the following formulation[4]:

$$t = t_0 + AV^n$$

Where $t$ is the travel time at flow $V$, $t_0$ is the free flow travel time, $n$ is the power of the flow delay curve, $A$ is a derived parameter based on the user defined free flow speed $S_0$ and speed at the link capacity $S_2$. The parameters applied are shown in **Table 3**, derived from COBA (Highways' Agency,

---

[3] Note that we haven't optimized the hyperparameters for all relevant baseline models. The model parameters from our core D-CLSTM-t model are applied to these models where relevant.
[4] For speed prediction by the speed flow curve, we assume the hourly flow is aggregated from the total flows in past four time steps. A passenger car unit(pcu) factor of 2.3 is applied to convert HGV from vehicles into pcus.

2002) in the UK.

| Description | $S_0$ (kph) | $S_2$ (kph) | Capacity(pcus) | n |
|---|---|---|---|---|
| Rural Motorway D4 | 113 | 81 | 9320 | 2.8 |
| Rural Motorway D3 | 113 | 81 | 6990 | 2.8 |

**Table 3 Speed flow curve parameters for D3 and D4 motorways in the UK**

Finally, we also develop test j and k models which are with the same model settings as test a. But the input data only includes historical speed for test j and flows for test k.

**Table 4** shows the model quantitative outputs measured by MSE and MAE on the validation datasets. A summary of findings are as follows:

- It can be seen that in general our core D-CLSTM-t model of test a achieves better performance than other models or methods in terms of prediction accuracy. The model performance of test b with traditional convolutional layers is close to test a, however, it took more than twice of time to train the model due to the increased hyperparameters in the model with traditional convolutional layers as mentioned earlier in this paper,
- The models with Dual CNN+LSTM frameworks generally perform better than a single CNN+LSTM (test c and d), which demonstrates a deeper neural network architecture can capture non-linearity of spatio-temporal patterns of traffic data. Meanwhile, the models with LSTM frameworks performs much better than CNN based model (test e).
- The results between test c and d show that the model with CNN-LSTM architecture with spatial information as the sequence performs significantly better than the model with temporal as sequence. The finding is consistent with the earlier works (Wu and Tan, 2016, Yu et al, 2017), which indicate that traffic prediction is more dependent on spatial closeness than the historical temporal features. In most cases, transient queues as a result of motorway shockwaves can be propagated immediately to upstream and dissipated after a short period. As such, speed prediction for upstream locations are heavily influenced by the downstream travel condition. Meanwhile, the implementation of LSTM on time axis means the site-specific speed information is largely lost which may also affect the accuracy of speed prediction.
- A comparison between test a and test d reveals that the incorporation of time of day and day of week in the model training benefits model performance with reduced MAE. This indicates that the short-term temporal variations and long-term periodicities are both important explanatory factors for travel speed prediction.
- The tree-based Light GBM, common-sense and speed-flow curve-based models generally perform worse than other CNN and LSTM based models. This shows these models fail to capture the underline traffic features which have strong correlations both in space and temporal domain. Surprisingly, the model with speed flow curve, as a common form of cost function in most strategic transport models, produces worst outcomes in terms of both MSE and MAE. This indicates that static and tree-based models fail to capture dynamic underlying travel patterns and has less predictive power than deep learning models.
- Finally, the model comparison between test a, j and k shows that a combination of speed and flow as model inputs for test a achieves the best performance against the other two models which only consider speed or flow as inputs. This indicates that the D-CLSTM-t is capable of capturing non-linear relationship between flow and speed which is critical for short-term speed prediction.

| Models | Model name | Model Description | MSE | MAE |
|---|---|---|---|---|
| test a | D-CLSTM-t | Dual CNN+LSTM with time marker, SeparableConv2D | 0.00138 | 0.0195 |
| test b | D-CLSTM-t, Conv2D | Dual CNN+LSTM with time marker, Conv2D | 0.00136 | 0.0197 |
| test c | CLSTM(S)-t | Single CNN+LSTM with time marker, SeparableConv2D | 0.00138 | 0.0200 |
| test d | CLSTM(T)-t | Single CNN+LSTM with time marker, SeparableConv2D | 0.00190 | 0.0235 |
| test e | D_CLSTM | Dual CNN+LSTM without time marker, SeparableConv2D | 0.00139 | 0.0205 |
| test f | CNN-t | Pure CNN with time marker | 0.00185 | 0.0277 |
| test g | LGBM | Light GBM | 0.00197 | 0.0234 |
| test h | Common-sense | assuming current speed as for the prediction in next 15 minutes | 0.00303 | 0.0265 |
| test i | UK_SFC | Using motorway speed curves | 0.01478 | 0.0739 |
| test j | D-CLSTM-t | The same as test a, model inputs only include historical speed data | 0.00169 | 0.0220 |
| test k | D-CLSTM-t | The same as test a, model inputs only include historical flow data | 0.00288 | 0.0287 |

**Table 4 model output comparisons on validation datasets**

## 6.2 D-CLSTM-t model performance and analysis

To examine model performance in terms of forecasting time window, a series of model tests were undertaken with the predicting window from the next 15 minutes up to the next 90 minutes by using D-LSTM-t model with setting identical to test a. The results are shown in **Figure 8(a)**. It can be seen that, as expected, the predictive power is less precise as the prediction window increases, with a MAE of 0.45 reported for speed prediction in next 90 minutes on the validation dataset. This indicates that the long-term temporal dependency is less important for the robustness of travel speed prediction. **Figure 8(b)** shows model prediction based on historical time windows from past 30 minutes up to 180 minutes to predict speed in next 15 minutes. As our data input is produced in every 15 minutes, the historical time window of 180 minutes would require a total data points of 12. The figure indicates that the best historical traffic information is around 60 minutes. Historical time window less than the range may not provide adequate information for accurate speed prediction, whilst any input data with longer historical time step may introduce extra noise to the model which may affect model performance.

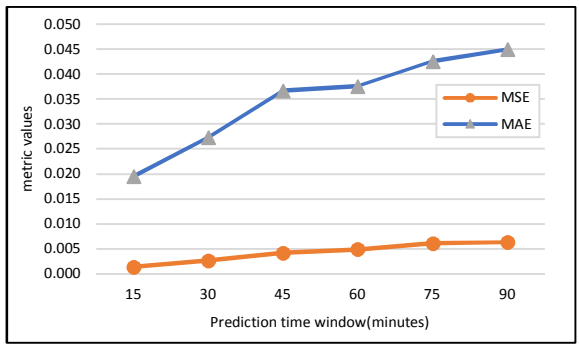 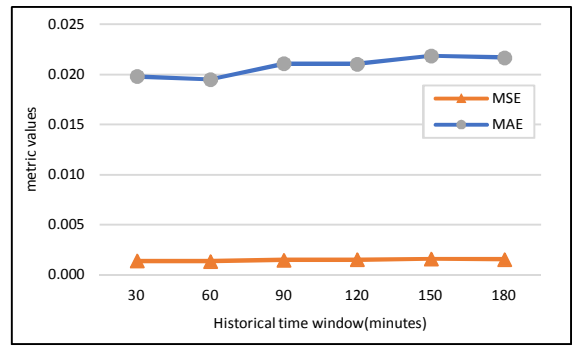

(a) Model outputs by prediction time window from 15 to 90 minutes

(b) Model outputs using data with historical time windows from 30 to 180 minutes

**Figure 8 Model output comparison by prediction time window and historical time window**

**Figure 9(a)** and **Figure 9(b)** show the two snapshots of speed prediction with residuals at site 4752A (between M25 J10 to J11) and 4717A (at M25 J10 CW off-slip road) on 24/09/2017. Note that the data on the day was used for validation during model training. Some observations of the two figures are shown as follows:

- Overall, the model captures travel time of the day patterns very well, which predict the starting and ending profile for both morning and evening peak. The model is struggling to make precise speed predictions during the peak periods. However, the residuals are less correlated and likely follow a normal distribution, which indicates the model leverages all available information in the training data with no significantly uncaptured leaking explanatory information into the residuals.
- Both pictures show that there are some noticeable differences for the speed prediction at the start of the day. This is probably due to the fact the model is trained in batch mode and there is no historical time-stepped data for sequence learning in LSTM.

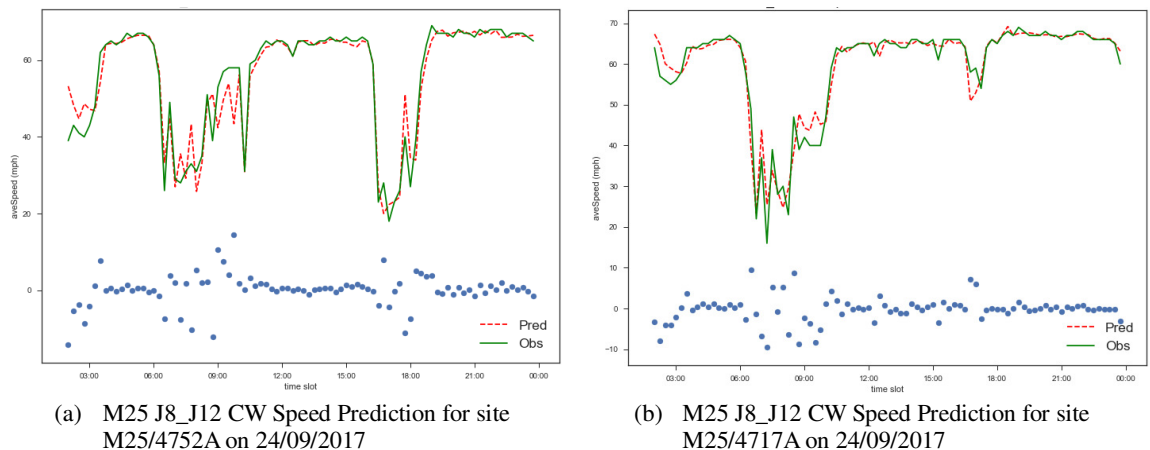

(a) M25 J8_J12 CW Speed Prediction for site M25/4752A on 24/09/2017

(b) M25 J8_J12 CW Speed Prediction for site M25/4717A on 24/09/2017

**Figure 9 Speed prediction snapshots on 24/09/2017**

### 6.3 Speed prediction under incident

To examine the model performance in terms of its speed prediction under unexpected accident. We took a snap shot of the validation dataset which hasn't been used for model training from 07:15 to 08:15AM on 04/09/2017. We simulate an accident taking place from 08:00 to 08:15 AM at site 8 (site ref:M25/4797A) and 9 (site ref:M25/4792A), which are located close to M25 J11 clockwise off-slip road. We assume when the accident happens, the total flow throughput and speed are reduced to 220 vehicles and 5 mph respectively.

**Figure 10** shows the speed prediction shown in red dash line in the case of the simulated accident for all 60 locations in next 15 minutes with a time ending at 08:30 AM, along with the observed speed sown in blue and the original speed prediction shown in green. It can be seen the model predict significantly reduced speed from the accident location to site 20 (site ref: M25/4727A) near M25 clockwise off-slip road at J10, with a length of approximate 6 kilometers. This implies a queue propagation speed around 24km/hour, which is consistent with the finding described from TRL's report (Notley et al., 2009). Interestingly, the model also predicts an increased speed at the downstream of the accident site, which appears sensible since the flows passing through these locations become light due to the bottleneck at the upstream accident site.

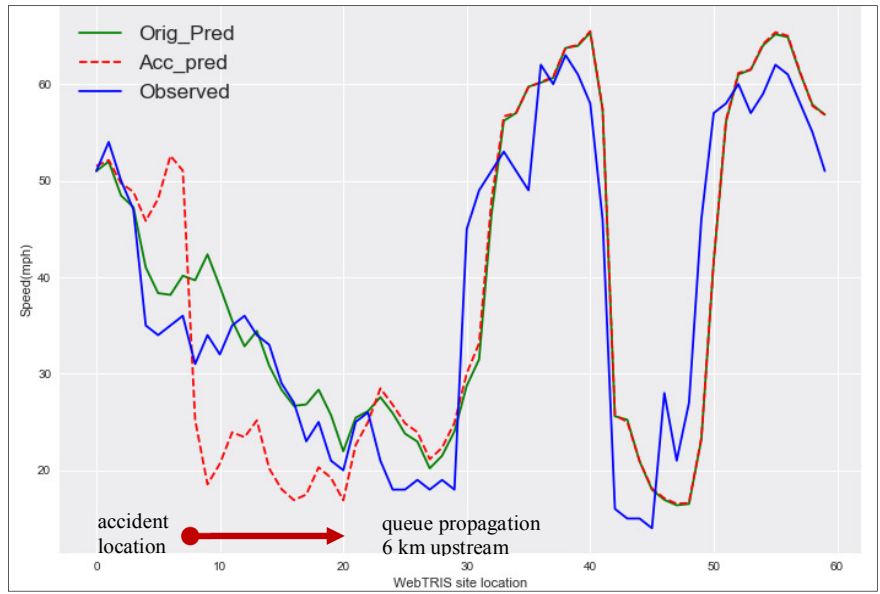

**Figure 10 speed on M25 in case of simulated accident between 08:15 to 08:30 AM, 04/09/2017.**

## 7.    Conclusion and Future Work

In this paper, we propose a data-driven modelling approach with a novel hierarchical D-CLSTM-t deep learning model for short-term traffic speed prediction, a framework combined with convolutional neural network (CNN) and long short-term model (LSTM) using historical flow and speed data. This proposed model structure applies two CNN models for spatio-temporal feature learning and extraction and LSTM model to capture characteristics in each time step. To capture flow temporal variations, time of the day and day of the week indicators are fused with trained features from CNN-LSTM. The model is trained end-to-end to predict travel speed robustly in 15 to 90 minutes in the future. Experiment results show that the D-CLSTM-t outperforms other baseline models considerably. Model tests show that speed prediction at upstream also responds sensibly to a sudden accident occurrence at downstream when flow throughput is significantly reduced. Our D-CLSTM-t model framework is also highly scalable for future extension such as for network-wide traffic prediction, which can also be improved by including additional features such as weather, long term seasonality and accident information.

During our model development and test, several issues have been noticed. First, for the models presented we have produced the image tensor inputs with site locations arranged sequential with the most downstream site on the top of the vertical axis. A sensitive test by reversing the location order shows similar model performance. To some extent, this demonstrates the limitation of location-invariant for the CNN+LSTM model architecture. In order words, the model makes no difference to the congestions occurring at upstream or downstream. It is thus ineffective to use convolution which uses a location-invariant filter to represent location-variant traffic data relationship, in both time and space domain. Inspired by the study (Lin et al., 2018) our test by incorporating the site-specific information as a Conv1D layer doesn't achieve better model performance, which we would like to explore further. Second, our training data was downloaded from the WebTRIS website, which has already been aggregated and pre-processed, which means that underlying patterns may have already been smoothed which cannot be learnt by the model. We also notice there are a few outliers of the data source. For example, there are some data points with zero flows but a wide range of speed values. This could be attributed to the flow free travel in the off-peak but also could be data errors, which could affect our model performance. Finally, further investigation is needed for a better understand of CNN-LSTM framework, and to make it more adaptable and robust for the unique data patterns in transport sector.

Our current model structure is developed for traffic prediction on road corridors where the input data can be transformed into time and space graphs nicely. This works well for the M25 case study due to the simplicity motorway network, but it is a challenge to extend the model framework network-wide in the urban context. One of the key tasks is to develop a 3-dimensional graph to present the 2-D location space and a time channel. The pioneering work done by Ma et al. (2017) could provide a possible solution.